\documentclass[runningheads]{llncs}
\usepackage{graphicx}
\usepackage[misc,geometry]{ifsym}
\begin{document}
\title{Explainable Deep Learning-based Solar Flare Prediction with post hoc Attention for Operational Forecasting}
%
% \author{Chetraj Pandey\Letter \orcidID{0000-0002-4699-4050} \and Rafal A. Angryk\orcidID{0000-0001-9598-8207} \and
% Berkay Aydin\orcidID{0000-0002-9799-9265}}
\author{Chetraj Pandey\Letter \inst{1} \orcidID{0000-0002-4699-4050} \and
Rafal A. Angryk \inst{1}\orcidID{0000-0001-9598-8207} \and
Manolis K. Georgoulis \inst{2}\orcidID{0000-0001-6913-1330} \and
Berkay Aydin \inst{1}\orcidID{0000-0002-9799-9265}}
\authorrunning{F. Author et al.}
\institute{Georgia State University, Atlanta, GA, USA \\
\email{\{cpandey1, rangryk, baydin2\}@gsu.edu} \\ \and 
Research Center for Astronomy and Applied Mathematics, Academy of Athens, Athens, Greece\\
\email{manolis.georgoulis@academyofathens.gr}}
%
% \author{Chetraj Pandey\inst{1,\dagger}\and
% Rafal A. Angryk\inst{1}\and
% Manolis K. Georgoulis\inst{2}\and
% Berkay Aydin\inst{1}}
%
\authorrunning{C. Pandey et al.}
% First names are 1bbreviated in the running head.
% If there are mor1 than two authors, 'et al.' is used.
%
% \institute{Georgia State University, Atlanta, GA, USA \and Research Center for Astronomy and Applied Mathematics, Academy of Athens, Athens, Greece\\
% \email{Correspondence$^\dagger$: cpandey1@gsu.edu}}
\titlerunning{Towards Explaining Solar Flare Prediction Model}
\maketitle              % typeset the header of the contribution
\begin{abstract}
This paper presents a post hoc analysis of a deep learning-based full-disk solar flare prediction model. We used hourly full-disk line-of-sight magnetogram images and selected binary prediction mode to predict the occurrence of $\geq$M1.0-class flares within 24 hours. We leveraged custom data augmentation and sample weighting to counter the inherent class-imbalance problem and used true skill statistic and Heidke skill score as evaluation metrics. Recent advancements in gradient-based attention methods allow us to interpret models by sending gradient signals to assign the burden of the decision on the input features. We interpret our model using three post hoc attention methods: (i) Guided Gradient-weighted Class Activation Mapping, (ii) Deep Shapley Additive Explanations, and (iii) Integrated Gradients. Our analysis shows that full-disk predictions of solar flares align with characteristics related to the active regions. The key findings of this study are: (1) We demonstrate that our full disk model can tangibly locate and predict near-limb solar flares, which is a critical feature for operational flare forecasting, (2) Our candidate model achieves an average TSS=0.51$\pm$0.05 and HSS=0.38$\pm$0.08, and (3) Our evaluation suggests that these models can learn conspicuous features corresponding to active regions from full-disk magnetograms.

\end{abstract}
\section{Introduction}
Solar flares are transient solar events of central importance to space weather forecasting, manifested as the sudden large eruption of electromagnetic radiation on the outermost atmosphere of the Sun. They are classified according to their peak X-ray flux level into the following five categories by National Oceanic and Atmospheric Administration (NOAA):  X $(\geq10^{-4}Wm^{-2})$, M $(\geq10^{-5}Wm^{-2})$, C $(\geq10^{-6}Wm^{-2})$, B $(\geq10^{-7}Wm^{-2})$, and A $(\geq10^{-8}Wm^{-2})$, where,  X$>$M$>$C$>$B$>$A \cite{spaceweather}. These flare classes are on a logarithmic scale, meaning that each class represents a tenfold increase in X-ray flux compared to the previous class. Large flares (M- and X-class) are scarce events that are more likely to incur a terrestrial impact and, therefore, the classes of interest that gather the attention of researchers. These flares may potentially disrupt the electricity supply chain, airline industry, and satellite communications, and pose radiation hazards to astronauts in space. To mitigate these risks, the necessity of a precise and reliable flare prediction model becomes imperative. 

Active regions (ARs) on the Sun are places characterized by the largest accumulations of dipolar magnetic flux in the solar atmosphere. Most operational flare forecasts target these regions of interest and issue predictions for individual ARs, which are the main initiators of space weather events. To issue a full-disk forecast with an AR-based model, the output flare probabilities for each active region are usually aggregated using a heuristic function as mentioned in \cite{Pandey2022f}. The heuristic function used to aggregate the final forecast operates under the assumption of conditional independence among ARs and that all ARs contribute equally to the aggregate forecast. This uniform weighting scheme may not accurately reflect the true influence of each AR on full-disk flare prediction probability. It is important to highlight that the weights of these ARs are generally unknown; there are no established methods to accurately determine them, nor are there any prior assumptions that guide the assignment of these weights.

Furthermore, the magnetic field measurements, employed by the AR-based forecasting techniques, are susceptible to severe projection effects as ARs get closer to limbs (to the degree that after $\pm$60$^{\circ}$ the magnetic field readings are distorted \cite{Falconer2016}); therefore, the aggregated full-disk flare probability is in fact, restrictive (i.e., from ARs in central locations) as the data in itself is limited to ARs located within $\pm$45$^{\circ}$  \cite{Li2020} to $\pm$70$^{\circ}$ \cite{Ji2020} and in some cases, even $\pm$30$^{\circ}$ \cite{Huang2018} due to severe projection effects \cite{Hoeksema2014}. As AR-based models include data up to $\pm$70$^{\circ}$, in the context of this paper, this upper limit ($\pm$70$^{\circ}$) is used as a boundary for central (within $\pm$70$^{\circ}$)  and near-limb regions (beyond $\pm$70$^{\circ}$).

In contrast to AR-based models, which use individual AR data from central locations, full-disk models use complete magnetogram images corresponding to the entire disk. These images are typically compressed JP2 (JPEG 2000) 8-bit representations (i.e., pixel values ranging from 0 to 255) derived from original magnetogram rasters which contain magnetic field strength values ranging from $\sim$$\pm$4500G. The compressed magnetogram images are used for shape-based parameters, e.g., size, directionality, borders, and inversion lines. Although projection effects still prevail in these images, full-disk models can learn from the near-limb areas. Thus, incorporating a full-disk model is essential to supplement AR-based models, enabling the prediction of flares in the Sun's near-limb areas and enhancing operational flare forecasting systems.
% Therefore, a full-disk model is appropriate to complement the AR-based counterparts as they can predict the flares appearing on the near-limb regions of the Sun and add a crucial element to the operational systems.

With recent advancements in machine learning and deep learning methods, their application in predicting solar flares has demonstrated great experimental success and accelerated the efforts of many interdisciplinary researchers \cite{Huang2018,Li2020,Nishizuka2018,Nishizuka_2017,Pandey2021,Pandey2022,pandey2023,Whitman2022}. Although deep learning methods have significantly enhanced solutions to image classification and computer vision problems, these models learn highly complex data representations, rendering them as black-box models. Consequently, the decision-making process within these models remains obscured, presenting a critical challenge for operational forecasting communities that rely on transparency to make informed decisions. Recently several empirical methods have been developed to explain and interpret the decisions made by deep neural networks. These are post hoc analysis methods (attribution methods) \cite{Linardatos2020}, meaning they focus on the analysis of trained models and do not contribute to the models' parameters while training. In this work, we primarily focus on developing a CNN-based full-disk model for solar flare prediction of $\geq$M1.0-class flares and evaluate and explain our model's performance by using three of the attribution methods: (i) Guided Gradient-weighted Class Activation Mapping (Guided Grad-CAM) \cite{gradcam}, (ii) Deep Shapley Additive Explanations (Deep SHAP) \cite{DeepShap}, and (iii) Integrated Gradients (IG) \cite{IntGrad}. More specifically, we show that our model's decisions are based on the characteristics corresponding to ARs, and our models can tackle the flares appearing on near-limb regions.

The rest of this paper is organized as follows. In Sec.~\ref{sec:rel}, we present the related work on flare forecasting. In Sec.~\ref{sec:data}, we present our methodology with data preparation and model architecture. In Sec.~\ref{sec:explain} we provide the description of all three post hoc explanation methods utilized in this work. In Sec.~\ref{sec:eval}, we present our experimental settings, and model evaluation, and discuss the interpretation of our models, and in Sec.~\ref{sec:cf}, we present our conclusions and future work.

\section{Related Work}\label{sec:rel}

There have been several attempts to predict solar flares using machine learning and deep learning models. A multi-layer perceptron-based model was applied for solar flare prediction of $\geq$C1.0- and $\geq$M1.0-class flares in \cite{Nishizuka2018} by utilizing 79 manually selected physical precursors extracted from multi-modal solar observations. A CNN-based flare forecasting model trained with solar AR patches extracted from line-of-sight (LoS) magnetograms within $\pm$30$^{\circ}$ of the central meridian to predict $\geq$C1.0-, $\geq$M1.0-, and $\geq$X1.0-class flares was presented in \cite{Huang2018}. Similarly, \cite{Li2020} also used a CNN-based model to issue binary class predictions for both $\geq$C1.0- and $\geq$M1.0-class flares within 24 hours using AR patches located within $\pm45^{\circ}$ of the central meridian. It is important to note that both of these models \cite{Huang2018}, \cite{Li2020}  are limited to a small portion of the observable disk in central locations (within $\pm30^{\circ}$ to $\pm45^{\circ}$) and thus possess the limited operational capability. 

% Moreover, a hybrid model combining GoogleLeNet \cite{Szegedy2015} and DenseNet \cite{Huang2017} was presented in \cite{Park2018}. They trained this model with a large volume of data from both HMI magnetograms, as well as magnetograms from Michelson Doppler Imager (MDI) onboard Solar and Heliospheric Observatory (SOHO), the predecessor of HMI/SDO to predict the occurrence of a $\geq$C-class within the next 24 hours. It should be noted that these two instruments are not yet cross-calibrated for use in forecasting, which might result in the discovery of deceptive or inadequate patterns. 

More recently, we presented a deep learning-based binary full-disk flare prediction model to predict $\geq$M1.0-class flares in \cite{Pandey2021} and to predict $\geq$C4.0- and $\geq$M1.0-class flares in \cite{Pandey2022} using bi-daily observations (i.e., two magnetograms per day) of full-disk LoS magnetograms. It is important to note that in \cite{Pandey2022} all the instances that fall between the $\geq$C4.0- and $\geq$M1.0-class flares were excluded in both training and validation sets. These particular sets of instances lie on the border of two binary outcomes and can be considered the harder-to-predict instances. These models are still black-box and do not provide explanations on any global and local variable importance. These explanations are important to understand the capabilities of full-disk models in near-limb regions and improve their trustworthiness in operational settings. In solar flare prediction, \cite{Bhattacharjee2020} used an occlusion-based method to interpret a CNN-based solar flare prediction model trained with AR patches. Similarly, \cite{Yi2021} presented a deep learning-based flare prediction model for predicting C-, M-, and X-class flares and provided visual explanations using Grad-CAM \cite{gradcam}, and Guided Backpropagation \cite{gbackprop}. They used daily observations of solar full-disk LoS magnetograms at 00:00 UT, and their models show limitations for the near-limb flares. Moreover, in \cite{Sun2022}, DeepLIFT \cite{Deeplift} and IG \cite{IntGrad} were evaluated for explaining CNN-based flare prediction model trained using tracked AR patches within $\pm70^{\circ}$.

This paper presents a CNN-based model to predict $\geq$M1.0-class flares, trained with full-disk LoS magnetograms images. The novel contributions of this paper are as follows: (i) We show an improved overall performance of a full-disk solar flare prediction model,  (ii) We utilized contemporary attribution methods to explain and interpret the decisions of our deep learning model, and (iii) More importantly, we show that our models can predict solar flares appearing on difficult-to-predict near-limb regions of the Sun. 

\section{Data and Model}\label{sec:data}
\vspace{-10pt}
\begin{figure}[tbh!]
\centering
\includegraphics[width=0.85\linewidth ]{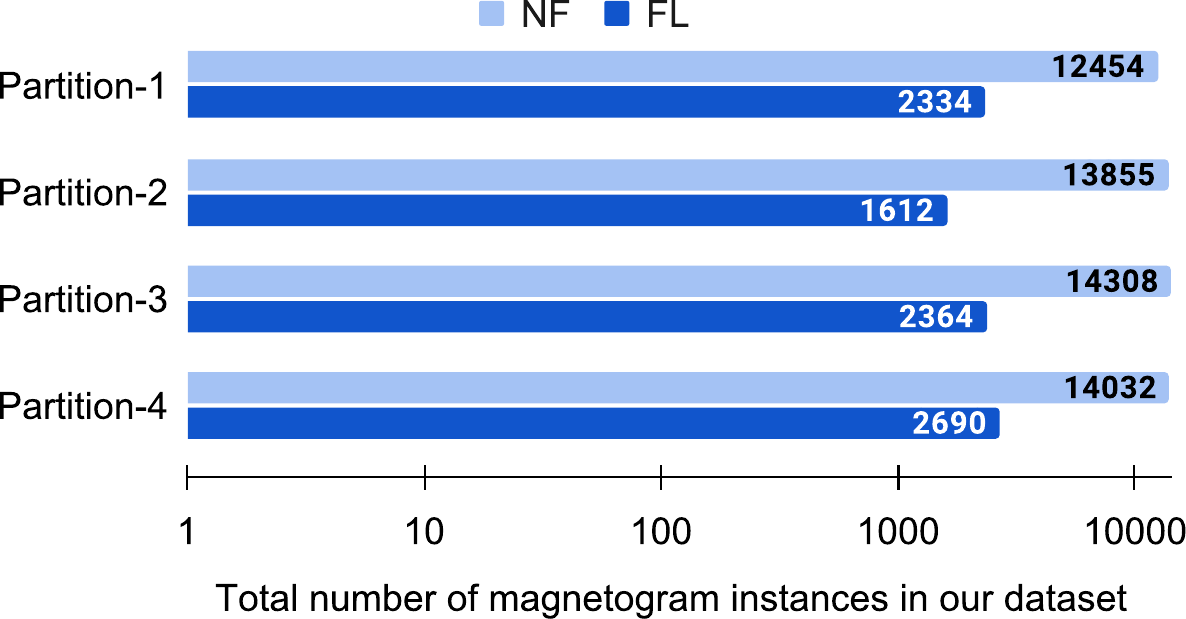}
\caption[]{Data distribution used in this study with four tri-monthly partitions for training $\geq$M1.0-class flare prediction models. Note: The length of the bars is in logarithmic scale.}
\label{fig:partitions}
\vspace{-10pt}
\end{figure}

We used full-disk LoS solar magnetograms obtained from the Helioseismic and Magnetic Imager (HMI) \cite{Schou2011} instrument onboard Solar Dynamics Observatory (SDO) \cite{Pesnell2011} available as compressed JP2 images in near real-time publicly via Helioviewer\footnote[1]{
Helioviewer: \url{https://api.helioviewer.org}}. To enhance computational efficiency for training the deep learning model, these compressed images are resized to a smaller resolution of 512x512 pixels. We sampled hourly instances of magnetogram images at [00:00, 01:00, ..., 23:00] each day from Dec 2010 to Dec 2018. We labeled our data with a prediction window of 24 hours. The images are labeled based on the maximum peak X-ray flux (converted to NOAA flare classes) within the next 24 hours. We collect a total of 63,649 images and label them such that if the maximum X-ray intensity of flare is weaker than M1.0, the observations are labeled as "No Flare" (NF: $<$M1.0) and $\geq$M1.0 ones are labeled as "Flare" (FL: $\geq$M1.0). This results in 54,649 instances for the NF-class and 9,000 instances (8,120 instances of M-class and 880 instances of X-class flares) for the FL-class. 

\begin{figure}[tbh!]
\vspace{-20pt}
  \centering
  \includegraphics[width=0.9\linewidth]{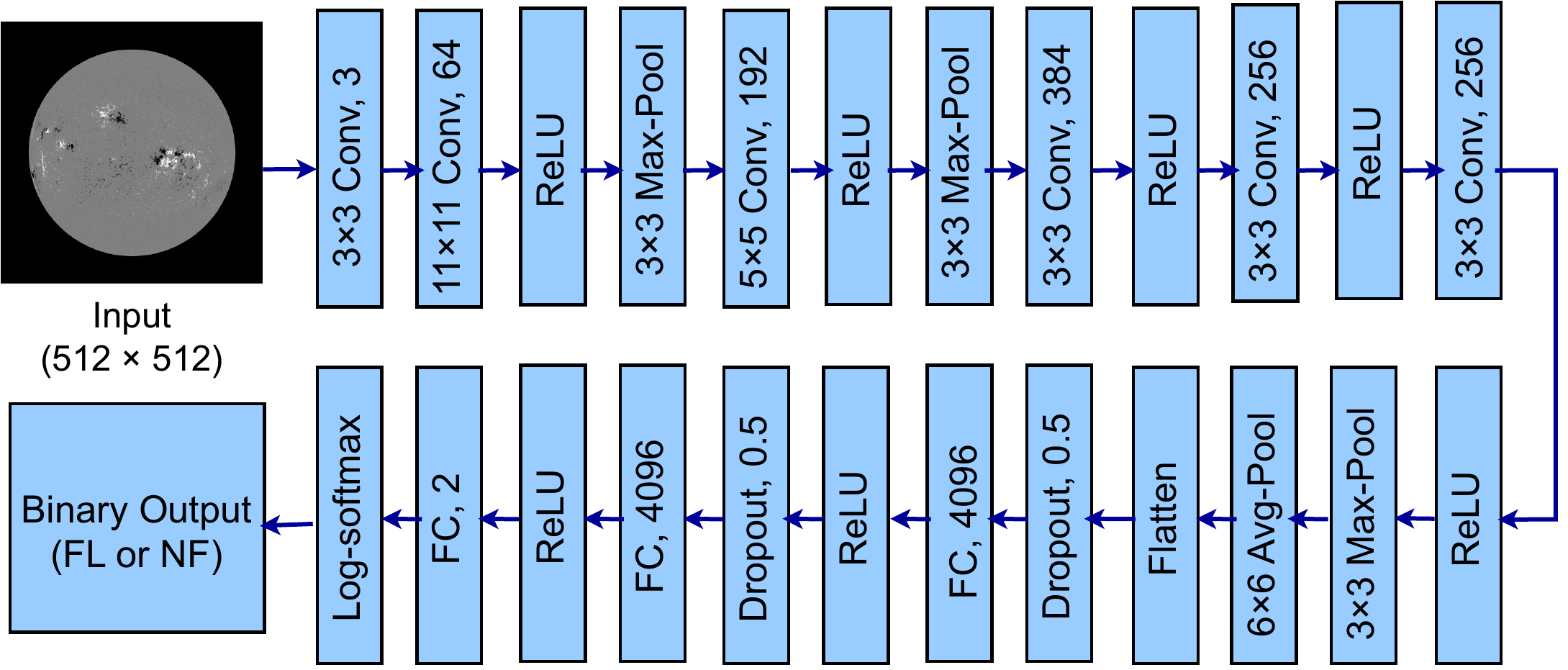}
  \caption{The architecture of our full-disk flare prediction model.}
  \label{fig:arch1}
  \vspace{-20pt}
\end{figure}

We finally split our data into four temporally non-overlapping tri-monthly partitions for the cross-validation experiments. This partitioning of the dataset is created by dividing the data timeline from Dec 2010 to Dec 2018 into four partitions, where Partition-1 contains data from Jan to Mar, Partition-2 contains data from Apr to Jun, Partition-3 contains data from Jul to Sep, and finally, Partition-4 contains data from Oct to Dec as shown in Fig.~\ref{fig:partitions}. As a result of the infrequent occurrence of $\geq$M1.0-class flares, the dataset exhibits a significant imbalance, with the ratio of FL to NF class being approximately 1:6.

In this work, we extend the AlexNet \cite{alex} model by concatenating a convolutional layer at the beginning of the network to make use of the pre-trained weights for our 1-channel input magnetogram images as the pre-trained model requires a 3-channel image as input to the network. Our added convolutional layer uses a 3$\times$3 kernel, size-1 stride, and outputs a 3-channel feature map which is then integrated into the standard AlexNet architecture as shown in Fig.~\ref{fig:arch1}. Furthermore, to efficiently utilize the pre-trained weights regardless of the architecture of the AlexNet model, which expects 224$\times$224, 3-channel image as input, we use the adaptive average pooling after feature extraction before the fully-connected layer to match the dimension on our 1-channel, 512$\times$512 magnetogram image. Overall, our model has six convolutional layers, three max-pool layers, one average-pool layer, and two fully-connected layers. 

\section{Interpretation Methods}\label{sec:explain}
Deep learning models are often deemed black-box due to their complex representations, resulting in interpretability, transparency, and consistency challenges concerning the patterns they learn \cite{Linardatos2020}. To address this, various methods \cite{Zhou2021} have been proposed to interpret CNNs. One common approach is using attribution methods, which visualize how specific parts of the input influence the model's decisions. Attribution methods generate attribution vectors (heat maps) representing the contribution of each input element to the model's decision. These methods can be perturbation-based (e.g., Local Interpretable Model-Agnostic Explanations (LIME) \cite{Ribeiro2016}), involving altering the input and measuring the difference in output, or gradient-based, calculating gradients via backpropagation to estimate attribution scores. While perturbation-based methods suffer from inconsistency issues due to creating Out-of-Distribution data  \cite{Qiu2022}, gradient-based methods are more robust to input perturbations and computationally efficient \cite{Nielsen2022}. Therefore, in this work, we employed three recent gradient-based methods to assess the interpretability of our models. By leveraging gradient-based techniques, known for their computational efficiency and robustness compared to perturbation-based methods, we aimed to visualize the decisions made by our model and gain insights into the specific characteristics in a magnetogram image that trigger the models' decisions. These methods allowed us to cross-validate and ensure the consistency of the explanations provided by our models, contributing to a more reliable and robust interpretation.\\

\noindent \textbf{Guided Grad-CAM: } The Guided Gradient-weighted Class Activation Mapping (Guided Grad-CAM) method \cite{gradcam} leverages the benefits of the Grad-CAM and guided backpropagation \cite{gbackprop}. Grad-CAM is a model-agnostic method that uses the class-specific gradient information flowing into the final convolutional layer of a CNN to produce a coarse localization map of the important regions in the image. Guided Backpropagation is based on the premise that the neurons act as detectors of certain image features, so it computes the gradient of the output with respect to the input, except that when propagating through ReLU functions, it only backpropagates the non-negative gradients and highlights the pixels that are important in the image. Attributions from Grad-CAM are class-discriminative and localize relevant image regions; however, do not highlight the fine-grained pixel importance as guided backpropagation \cite{Chattopadhay2018}. Guided Grad-CAM combines the fine-grained details of guided backpropagation with the course localization advantages of Grad-CAM and is computed as the element-wise product of guided backpropagation with the upsampled Grad-CAM attributions.\\

\noindent \textbf{Deep SHAP: } SHAP values (SHapley Additive exPlanations) \cite{DeepShap} is a method based on cooperative game theory\cite{1952} and used to increase the transparency and interpretability of machine learning models. SHAP shows the contribution of each feature to the prediction of the model, it does not evaluate the quality of the prediction itself. The contribution of each feature is calculated using cooperative game theory and Shapley values to assess how much each feature adds to the difference between the actual prediction and the average prediction. For deep-learning models, Deep SHAP \cite{DeepShap} is considered an enhanced version of the DeepLIFT algorithm  \cite{Deeplift}, where we approximate the conditional expectations of SHAP values using a selection of baseline samples from the dataset. The baselines typically contain a set of representative samples from the same distribution as the input data. For each input sample, it computes DeepLIFT attribution with respect to each baseline and averages resulting attributions. This method assumes that input features are independent of one another, and the explanations are modeled through the additive composition of feature effects. \\

\noindent \textbf{Integrated Gradients: }
The last method we will analyze in this study is Integrated Gradients (IG) \cite{IntGrad}, which quantifies feature attributions by integrating the gradients of the model's output along a straight-line path from a baseline reference to the input feature under consideration. This method requires an extra input as the baseline, representing the non-appearance of the feature in the original image which is typically an all-zero vector. IG is favored for its completeness property, where the sum of integrated gradients for all features precisely equals the difference between the model's output for the given input and the baseline input values. This property ensures that the feature attributions accurately represent each feature's individual contribution to the model output, allowing us to reliably recover the model's output value by summing these contributions \cite{Sturmfels2020}.

\section{Experimental Evaluation}\label{sec:eval}
\subsection{Experimental Settings}
We trained a full-disk flare prediction model with stochastic gradient descent (SGD) as an optimizer and negative log-likelihood (NLL) as the objective function. Our model is initialized with pre-trained weights of AlexNet Model \cite{alex}, and then we make use of a dynamic learning rate (initialized at 0.0099 and reduced 5$\%$) to further train the model to 40 epochs with a batch size of 64. We address the class-imbalance issue using data augmentation and class weights to the loss function. We use three augmentation techniques: vertical flipping, horizontal flipping, and +5$^{\circ}$ to -5$^{\circ}$ rotations. We augment the data for both classes (where the entire FL-class data are augmented three times with three augmentation techniques and NF-class is augmented once randomly). We then adjust class weights inversely proportional to the class frequencies after augmentations. The use of class weights penalizes the misclassification made in the minority class. Our models are trained as 4-fold cross-validation experiments with each fold representing a different partition serving as the test set. Specifically, Fold-1 corresponds to Partition-1, Fold-2 corresponds to Partition-2, and so on.

We evaluate the performance of our models using two widely-used forecast skills scores: True Skill Statistics (TSS, in Eq.~\ref{eq:TSS}) and Heidke Skill Score (HSS, in Eq.~\ref{eq:HSS}), derived from the elements of confusion matrix: True Positives (TP), True Negatives (TN), False Positives (FP), and False Negatives (FN). In the context of our paper, the FL class is the positive outcome and NF is the negative. 
\begin{equation}\label{eq:TSS}
    TSS = \frac{TP}{TP+FN} - \frac{FP}{FP+TN}
\end{equation}

\begin{equation}\label{eq:HSS}
    HSS = 2\times \frac{TP \times TN - FN \times FP}{((P \times (FN + TN) + (TP + FP) \times N))},
\end{equation}
\begin{center}
    where $N = TN + FP$ and $P = TP + FN$.
\end{center} 

\begin{equation}\label{eq:rec}
    Recall = \frac{TP}{TP+FN}
\end{equation}
TSS and HSS values range from -1 to 1, where 1 indicates all correct predictions, -1 represents all incorrect predictions, and 0 represents no skill. In contrast to TSS, HSS is an imbalance-aware metric, and it is common practice to use HSS in combination with TSS for the solar flare prediction models due to the high class-imbalance ratio present in the datasets. For a balanced dataset, these metrics are equivalent \cite{Ahmadzadeh2021}. In solar flare prediction, TSS and HSS are the preferred choices of evaluation metrics compared to commonly used metrics in image classification (e.g., accuracy) as they ensure a comprehensive and reliable evaluation of predictive capabilities, especially in scenarios with imbalanced class distributions. Lastly, we report the subclass and overall recall (shown in Eq.~\ref{eq:rec}) for flaring instances (M- and X-class) to assess the prediction sensitivity of our models in central and near-limb regions.  To reproduce this work, the source code and experimental results can be accessed from our open-source repository \cite{sourcecode}. 

\subsection{Model Evaluation}

Our models have on average TSS$\sim$0.51$\pm$0.05 and HSS$\sim$0.38$\pm$0.08, which improves over the performance of \cite{Pandey2021} by $\sim$4\% in terms of TSS (reported 0.47$\pm$0.06) and by $\sim$3\% in terms of HSS (reported 0.35$\pm$0.05) \footnote[2]{While there are several other works (mentioned in Sec.~\ref{sec:rel}) in solar flare prediction, the results of these models are not directly comparable since they employ different datasets, data timelines, and data partitioning strategies.}. The detailed experimental results for each fold are shown in Table.~\ref{table:fold-results}.  

\begin{table}[tbh!]
\setlength{\tabcolsep}{6pt}
\renewcommand{\arraystretch}{1.4}
\caption{A comprehensive overview of 4-fold cross-validation experiments, showing all the four outcomes of confusion matrices (TP, FP, TN, FN) evaluated on the test sets, and performance of our models in terms of two skill scores (TSS and HSS).}
\begin{center}
\begin{tabular}{cccccccc}
\hline
Folds & TP & FP & TN & FN & TSS & HSS \\
\hline
Fold-1 & 1,720 & 1,943 & 10,511 & 614 & 0.58 & 0.47 \\
Fold-2 & 1,155 & 3,083 & 10,772 & 457 & 0.49 & 0.29 \\
Fold-3 & 1,585 & 2,668 & 11,640 & 779 & 0.48 & 0.36 \\
Fold-4 & 1,706 & 2,241 & 11,791 & 984 & 0.47 & 0.40 \\
\hline
Aggregated & 6,166 & 9,935 & 44,714 & 2,834 & \textbf{0.51$\pm$0.05} & \textbf{0.38$\pm$0.08} \\
\hline
\end{tabular}
\end{center}
\label{table:fold-results}
\vspace{-5pt}
\end{table}

\begin{table}[tbh!]
\setlength{\tabcolsep}{3pt}
\renewcommand{\arraystretch}{1.4}
\caption{Counts of correctly (TP) and incorrectly (FN) classified X- and M-class flares in central ($|longitude|$$\leq\pm70^{\circ}$) and near-limb locations. The recall across different location groups is also presented. Counts are aggregated across folds.}

\begin{center}
 \begin{tabular}{r c c c c c c}
\hline
 & 
\multicolumn{3}{c}{Within $\pm$70$^{\circ}$} %%\\               %% <-- mistake
%% \cline{1-2}                               %% <-- mistake
&                                            %% <--  addition
\multicolumn{3}{c}{Beyond $\pm$70$^{\circ}$}\\
          %% <--  Changed
Flare-Class & TP  & FN  & Recall  & TP   &FN & Recall \\
\hline
%  &    &     &    & &    & \\
X-Class  &  637  & 31  & 0.95 & 157 & 55 & 0.74\\

M-Class &  4,229 & 1,601  & 0.73 & 1,143   &1,147 & 0.50\\

Total (X\&M) & 4,866 & 1,632 &0.75& 1,300 & 1,202 & 0.52\\ 
\hline
\end{tabular}
\end{center}
\label{table:comp}
\vspace{-5pt}
\end{table}

\begin{figure}[tbh!]
\centering
\includegraphics[width=0.93\linewidth ]{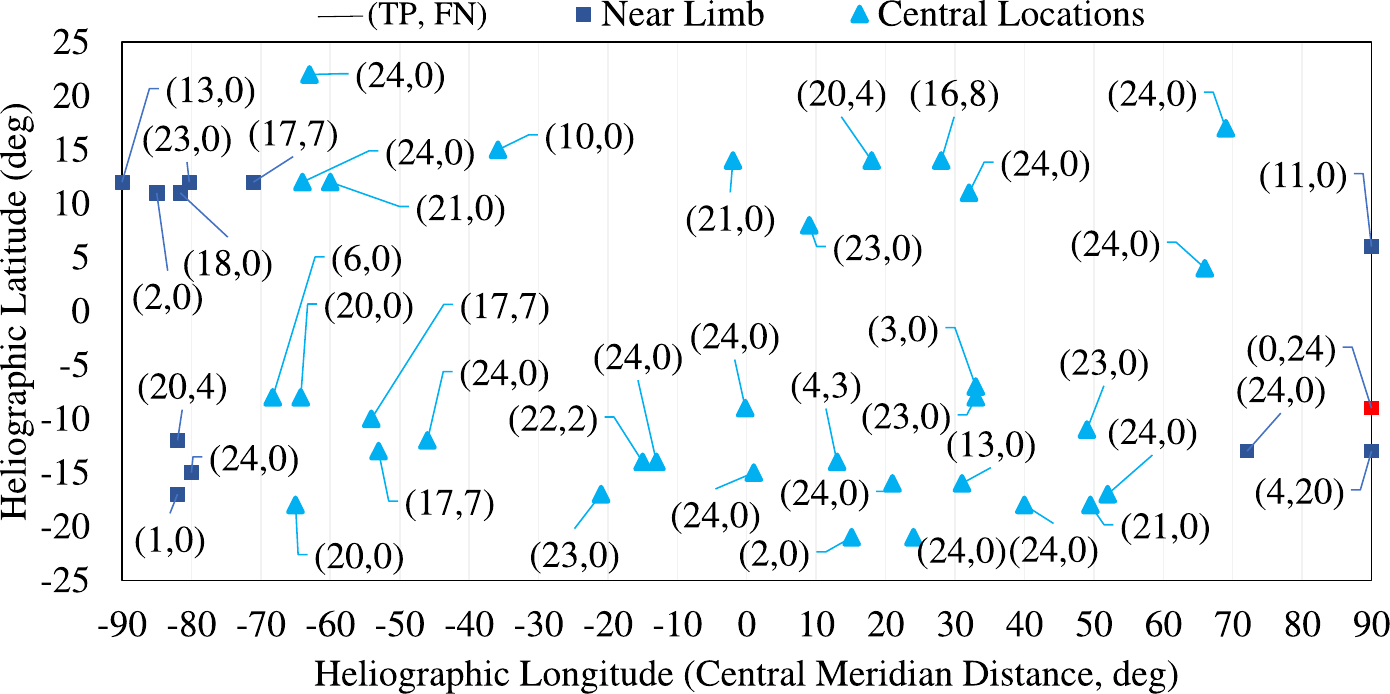}
\caption[]{A scatterplot to quantify the performance of our models in terms of True Positives (TP) and False Negatives (FN) for X-class flares grouped by flare locations. The flare events beyond $\pm$70$^{\circ}$ longitude are represented as near-limb events.  Note: (i) Red marker is for locations with zero TP. (ii) For some locations, TP$+$FN$<$24, given that we used hourly instances, is due to the unavailable instances from the source.}
\label{fig:xclass}
\vspace{-10pt}
\end{figure}

In addition, we evaluate our results for correctly predicted and missed flare counts for class-specific flares (X-class and M-class) in central locations (within $\pm$70$^{\circ}$) and near-limb locations (beyond $\pm$70$^{\circ}$) of the Sun as shown in Table \ref{table:comp}. We observe that our models made correct predictions for $\sim$95\% of the X-class flares and $\sim$73\% of the M-class flares in central locations. Similarly, our models show a compelling performance for flares appearing on near-limb locations of the Sun, where $\sim$74\% of the X-class and $\sim$50\%  of the M-class flares are predicted correctly. This is important because, to our knowledge, the prediction of near-limb flares is often overlooked. More false positives in M-class are expected because of the model's inability to distinguish bordering class flares (C4.0 to C9.9) from $\geq$M1.0-class flares, which we have observed empirically in our prior work \cite{Pandey2022} as well. Overall, we observed that $\sim$90\% and $\sim$66\% of the X-class and M-class flares, respectively, are predicted correctly by our models.

Furthermore, given that we sample our data with a 1-hour cadence resulting in 24 instances per day unless there are gaps due to unavailable data instances, any given flare instance is expected to be in the prediction window of 24 instances. X-class flares are relatively large flares that often dominate the prediction window. Therefore, we analyzed the predictions on X-class flares and observed that from a total of 45 X-class flare locations, our models correctly predict the occurrence of a flare at least once for 44 of them, as shown in Fig.~\ref{fig:xclass}. In particular, we show that the full-disk model presented in this paper can predict flares appearing on near-limb locations of the Sun at great accuracy, which provides a crucial addition to operational flare forecasting systems.

\subsection{Model Interpretation}\label{sec:dis}

\begin{figure}[bth!]
\centering
\includegraphics[width=0.9\linewidth ]{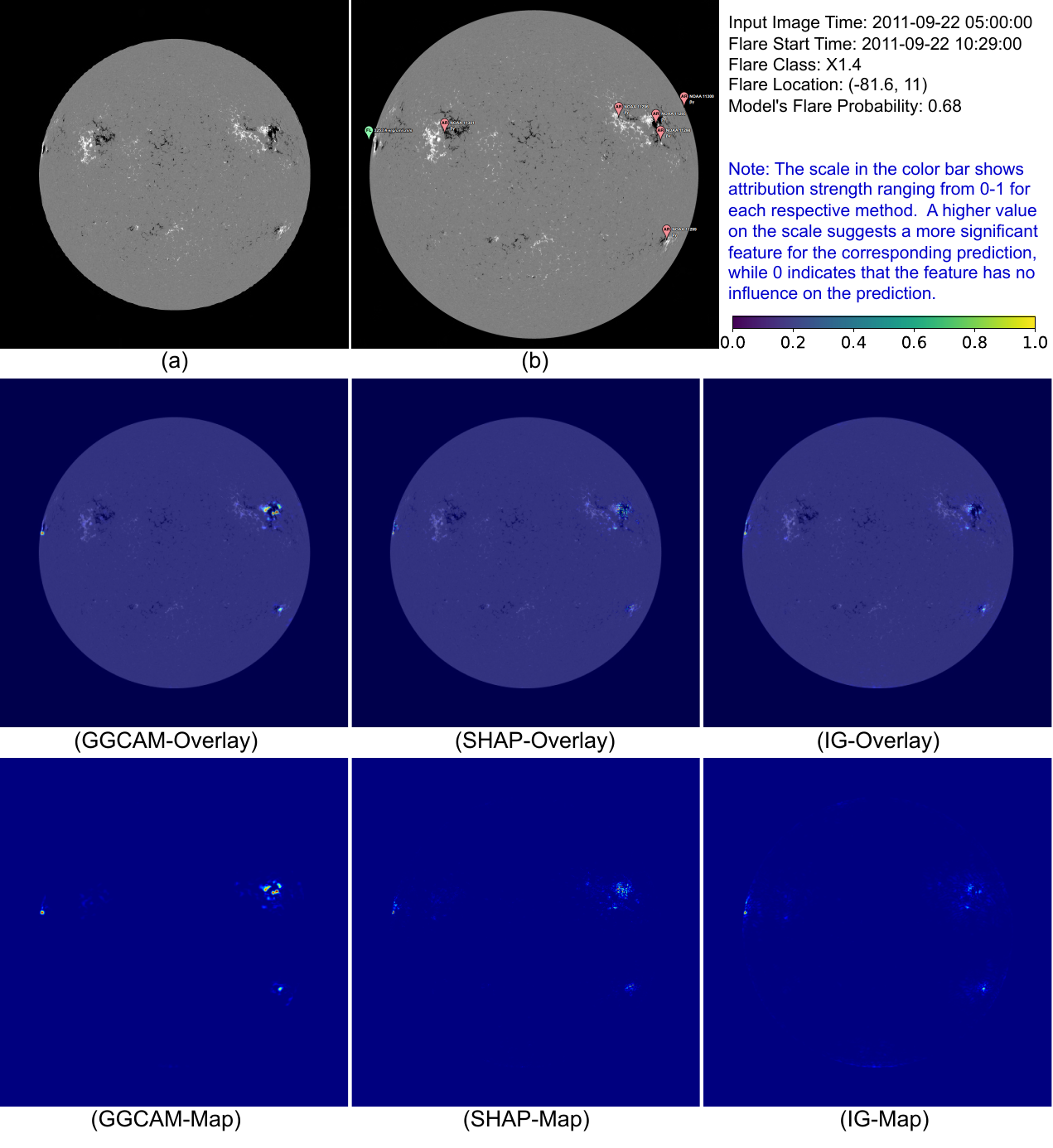}
\caption[]{A visual explanation for a correctly predicted near-limb FL-class instance. (a) Actual magnetogram from the dataset used as the input image. (b) Annotated full-disk magnetogram at flare start time, showing flare location (green flag) and NOAA ARs (red flags). Overlays (GGCAM, SHAP, IG) depict the input image overlayed with attributions, and Maps (GGCAM, SHAP, IG) showcase the attribution maps obtained from Guided Grad-CAM, Deep SHAP, and Integrated Gradients respectively.}
\label{fig:tp}
\vspace{-15pt}
\end{figure}

\begin{figure}[bth!]
\centering
\includegraphics[width=0.9\linewidth ]{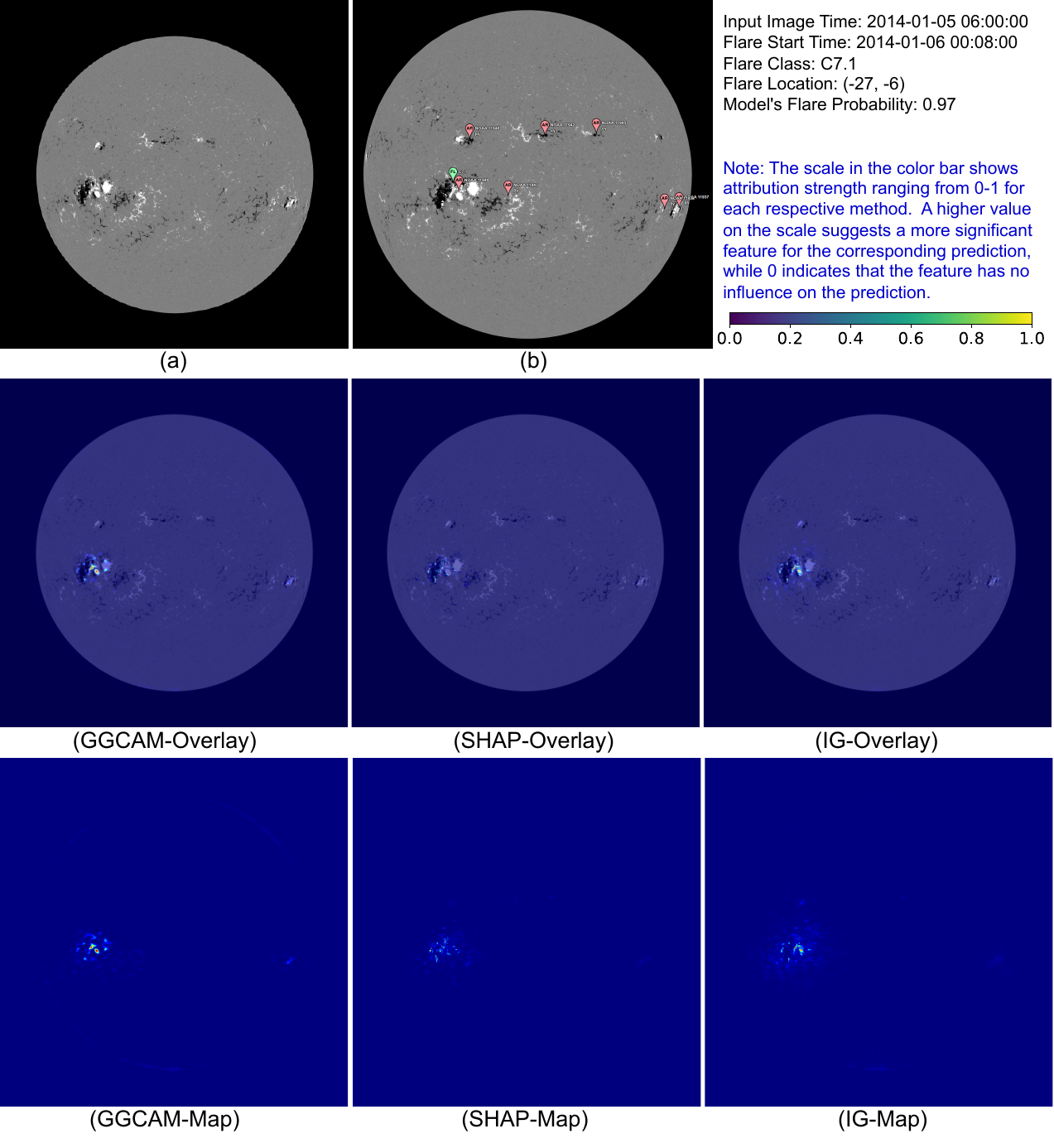}
\caption[]{A visual explanation for an incorrectly predicted NF-class instance. (a) Actual magnetogram from the dataset used as the input image. (b) Annotated full-disk magnetogram at flare start time, showing flare location (green flag) and NOAA ARs (red flags). Overlays (GGCAM, SHAP, IG) depict the input image overlayed with attributions, and Maps (GGCAM, SHAP, IG) showcase the attribution maps obtained from Guided Grad-CAM, Deep SHAP, and Integrated Gradients respectively.}
\label{fig:fp}
\vspace{-15pt}
\end{figure}

In this section, we present a case study, interpreting the visual explanations generated by our model, and also discuss the implications of these explanations in the operational forecasting scenario.  For this, we use the visualizations generated using all three post hoc explanation methods mentioned earlier in Sec.~\ref{sec:explain} for two instances: (i) a correctly predicted (TP) near-limb flare instance and (ii) an incorrectly predicted (FP) instance. 

Firstly, we interpret the predictions of our model for a correctly predicted X1.4-class flare observed on 2011-09-22 at 10:29:00 UTC on the East limb (note that East and West are reversed in solar coordinates). We generate a visual explanation using all three attribution methods. We utilized an input image from 2011-09-22 05:00:00 UTC (approximately 5.5 hours prior to the flare event) where the sunspot corresponding to the flare becomes visible in the magnetogram image. Interestingly, we observed that the pixels covering the AR on the East limb, which is responsible for the eventual X1.4 flare, are activated, as shown in Fig.~\ref{fig:tp}. Note that the location of the flare is indicated by a green flag and all visible NOAA ARs are indicated by red flags in Fig.~\ref{fig:tp} (b). The model focuses on specific ARs, including the relatively smaller AR on the East limb, even though other ARs are present in the magnetogram image. The visualization of attribution maps suggests that, for this particular prediction, the region responsible for the flare event is attributed as important, contributing to the consequent decision. This finding is consistent across all three methods, corroborating the explanation's reliability. However, Guided Grad-CAM and Deep SHAP provide finer details by suppressing noise compared to IG.

Similarly, to analyze a false positive case, we present an example of a C7.1 flare observed on 2014-01-06 at 00:08:00 UTC. To explain the result, we used an input magnetogram instance from 2014-01-05 06:00:00 UTC ($\sim$18 hours prior to the event). The model's prediction probability for this instance being an FL-class is $\sim$0.97. Therefore, we seek a visual explanation of this prediction using all three interpretation methods. Upon analysis, we observed that the prediction mainly relies on only one AR, which indeed corresponds to the location of the eventual C7.1 flare (indicated by the green flag) when visualized with all three attribution methods, as shown in Fig.~\ref{fig:fp}. This incorrect prediction can be attributed to the interference of the bordering class flares mentioned in \cite{Pandey2022}. Such interference poses a problem for binary flare prediction models. We noticed that out of 25,150 C-class flares, 9,240 flares led to incorrect predictions, accounting for approximately 37\% of the total C-class flares in our dataset.

These two examples, although not exhaustive, carry significant implications for operational forecasting systems. By incorporating visual explanations into the forecasting process, in addition to providing a full-disk flare prediction probability, we have the capability to identify potential flare event locations among all visible ARs precisely. This is invaluable for improving the accuracy and reliability of solar flare forecasts, aiding in effective risk assessment and mitigation strategies. Furthermore, it provides a deeper understanding of the underlying factors contributing to flare occurrences, empowering researchers and space weather experts to make more informed decisions and take timely actions to safeguard critical infrastructure and space assets.

\section{Conclusion and Future Work}\label{sec:cf}

In this work, we used three recent gradient-based methods to interpret the predictions of our AlexNet-based binary flare prediction model trained for the prediction of $\geq$M1.0-class flares. We addressed the highly overlooked problem of flares appearing in near-limb locations of the Sun, and our model shows a compelling performance for such events. Furthermore, we evaluated our model's predictions with visual explanations, showing that the decisions are primarily capturing characteristics corresponding to the active regions in the magnetogram instance. Although our model shows improved capability, still suffers from high false positives attributed to high C-class flares. As an extension, we plan to study the individual class characteristics to obtain a better way of segregating these flare classes considering the background flux and generate a new set of labels that can better address the issue with border class flares. Furthermore, at this point, the models are only looking at the spatial patterns in our data, and we intend to widen this work toward spatiotemporal models to improve the performance.\\

\noindent\textbf{Acknowledgements: }
This project is supported in part under two NSF awards \#2104004 and \#1931555, jointly by the Office of Advanced Cyberinfrastructure within the Directorate for Computer and Information Science and Engineering, the Division of Astronomical Sciences within the Directorate for Mathematical and Physical Sciences, and the Solar Terrestrial Physics Program and the Division of Integrative and Collaborative Education and Research within the Directorate for Geosciences. This work is also partially supported by the National Aeronautics and Space Administration (NASA) grant award \#80NSSC22K0272. Data used in this study is a courtesy of NASA/SDO and the AIA, EVE, and HMI science teams and NOAA National Geophysical Data Center (NGDC).

\bibliographystyle{splncs04}
\bibliography{bibfile}
\end{document}